\documentclass[runningheads]{llncs}
\usepackage[T1]{fontenc}
\usepackage{graphicx}
\usepackage{xcolor}
\usepackage{amsmath}
\usepackage{esvect}
\usepackage{amssymb}
\usepackage{multirow}
\usepackage{diagbox}
\usepackage{textcomp}
\usepackage{booktabs}
\usepackage{algorithm}
\usepackage{algpseudocode}

\renewcommand{\arraystretch}{1.2}
\DeclareMathOperator*{\concat}{%
    \mathchoice%
        {\Big\Vert}%
        {\big\Vert}%
        {\Vert}%
        {\Vert}%
}
\begin{document}
\title{Attention Maps in 3D Shape Classification for Dental Stage Estimation with Class Node Graph Attention Networks}
\titlerunning{Class Node Graph Attention Networks}

\author{Barkin Buyukcakir\inst{1}\orcidID{0000-0001-5694-6100} \and 
Rocharles Cavalcante Fontenele\inst{2,3}\orcidID{0000-0002-6426-9768} \and 
{Reinhilde Jacobs}\inst{2,3}\orcidID{0000-0002-3461-0363}
\and {Jannick De Tobel}\inst{4}\orcidID{0000-0002-8846-7339}
\and {Patrick Thevissen}\inst{5}\orcidID{0000-0003-0890-6264}
\and {Dirk Vandermeulen}\inst{1}\orcidID{0000-0002-4052-5296}
\and {Peter Claes}\inst{1,6}\orcidID{0000-0001-9489-9819}
}



%
\authorrunning{Buyukcakir et al.}
%

\institute{KU Leuven, Department of Electrical Engineering (ESAT) - Processing Speech and Images (PSI), Leuven, 3000, Belgium \\
\and KU Leuven, Department of Imaging \& Pathology, Oral and Maxillofacial Surgery - Imaging \& Pathology (OMFS-IMPATH), Leuven, 3000, 
\and KU Leuven, University Hospitals Leuven, Department of Oral \& Maxillofacial Surgery, Leuven, 3000, Belgium\\
\and Ghent University, Deparment of Diagnostic Sciences, Ghent, 9000, Belgium \\
\and KU Leuven, Department of Imaging \& Pathology, Forensic Odontology, Leuven, 3000, Belgium \\
\and KU Leuven, Department of Human Genetics, Leuven, 3000, Belgium
} 

\maketitle              
\begin{abstract}
Deep learning offers a promising avenue for automating many recognition tasks in fields such as medicine and forensics. However, the "black box" nature of these models hinders their adoption in high-stakes applications where trust and accountability are required. For 3D shape recognition tasks in particular, this paper introduces the Class Node Graph Attention Network (CGAT) architecture to address this need. Applied to 3D meshes of third molars derived from CBCT images, for Demirjian stage allocation, CGAT utilizes graph attention convolutions and an inherent attention mechanism, visualized via attention rollout, to explain its decision-making process. We evaluated the local mean curvature and distance to centroid node features, both individually and in combination, as well as model depth, finding that models incorporating directed edges to a global CLS node produced more intuitive attention maps, while also yielding desirable classification performance. We analyzed the attention-based explanations of the models, and their predictive performances to propose optimal settings for the CGAT. The combination of local mean curvature and distance to centroid as node features yielded a slight performance increase with a 0.76 weighted F1 score, and more comprehensive attention visualizations. The CGAT architecture's ability to generate human-understandable attention maps can enhance trust and facilitate expert validation of model decisions. While demonstrated on dental data, CGAT is broadly applicable to graph-based classification and regression tasks, promoting wider adoption of transparent and competitive deep learning models in high-stakes environments.

\end{abstract}
\section{Introduction}

The field of deep learning (DL) has witnessed remarkable advancements in recent years, leading to transformative capabilities across a multitude of domains, and becoming state-of-the-art in tasks such as image recognition, natural language processing, text generation, and increasingly, the analysis of 3D shapes. Despite these significant strides and a plethora of published studies showcasing their potential, the widespread adoption of DL methods in the practice of high-stakes applications, particularly in sectors like healthcare and forensics, remains surprisingly limited. A primary factor contributing to this cautious uptake is the inherent opaqueness of many cutting-edge DL architectures. The inability to fully understand or interpret the decision processes of these complex algorithms raises concern about the reliability, trustworthiness, and potential biases, especially when critical outcomes regarding human health and rights are of concern \cite{dhar2023challenges}. This underscores a pressing need for explainable artificial intelligence (XAI) methods that can provide transparency and insight into model behavior. With the recent global focus by organizations such as the U.S. Defense Advanced Research Projects Agency (DARPA) \cite{national2019national}, and the European Union with its General Data Protection Regulation (GDPR) \cite{sartor2020impact}, on interpretable deep learning systems intensifying, explainability in deep learning has become more important.

DL models have been applied extensively in the relevant literature on a host of medical imaging modalities, for diagnostic purposes, such as disease detection and diagnosis, segmentation of biological structures, and image registration \cite{piccialli2021survey}. In the forensic domain, DL algorithms have found use in the evaluation of biometrics, for authentication and identification tasks on various data such as faces and fingerprints \cite{minaee2023biometrics}. However, most of these applications have been based on two-dimensional (2D) data, where the primary means of explainability has been attention or saliency maps, where the most influential regions of the input images in model prediction are highlighted on the images as a form of decision explanation \cite{hassanin2024visual}. However, when 3D data is concerned, the direct applications of the tried and tested DL architectures were not immediately possible due to the inherent properties of 3D shape representations raising new requirements such as permutation invariance or incorporation of topology \cite{chen2019deep}, and as such, the direct equivalent of attention maps for 3D data has not yet been proposed. 

Geometric deep learning (GDL) subsequently emerged as a specialized subfield of DL, providing powerful frameworks for analyzing 3D structures. GDL models operate most commonly on one of four representations of 3D data, which are: (1) point clouds where the 3D shape is depicted as a collection of points in space, (2) meshes, which depict discrete surfaces defined by vertices and edges, highlighting the underlying topology of the shapes, and (3) graphs, which are abstract structures defined by a set of nodes with arbitrary features encoded in them, and the edges connecting these nodes together which represent some relation between them  \cite{gezawa2020review,cao2020comprehensive}. For point clouds, PointNet and PointNet++ have been the pioneering architectures due to their agnosticism to input permutation \cite{qi2017pointnet,qi2017pointnet++}. These models have been applied to human identification based on 3D teeth models \cite{liu2023human}, Alzheimer's disease diagnosis on hippocampal surfaces \cite{yang2024pre} and age prediction \cite{yang2024coordinate}. The main difficulty of point cloud representations is that due to the lack of geodesic connections, two points in close proximity can be valid neighbors even though they are topologically unassociated. This difficulty is alleviated by mesh and, more generally, graph representations, where connectivity is defined. Operating on meshes, methods such as MeshCNN \cite{hanocka2019meshcnn} and SpiralNet++ \cite{gong2019spiralnet++} have been introduced, and applied in tasks such as syndrome classification \cite{mahdi2022multi} and age prediction based on cortical surface shape \cite{vosylius2020geometric}. Although mesh-based approaches solve the topology problem in point clouds, many of them require mesh vertices to be in correspondence with each other across samples, thus requiring mesh registration as a preprocessing step. Graph neural networks (GNNs) do not have such requirements as they commonly feature a permutation-invariant convolution operation which operates solely on the features embedded in the nodes of input graphs. Some commonly used examples of GNN architectures are graph convolutional networks (GCNs) \cite{kipf2016semi}, GraphSAGE \cite{hamilton2017inductive} and graph attention networks (GATs) \cite{velickovic2017graph}, which have been employed in tumor segmentation \cite {mohammadi2024advancing}, age and sex prediction \cite{shehata2022comparative,besson2021geometric} based on anatomical scans, and facial landmarking \cite{facchi2025graph}. 

Despite this extensive literature on 3D shape analysis using geometric deep learning, widespread clinical use of these methods has remained limited, due to the lack of an explainability mechanism that can provide the basis for model decisions, with perhaps the exception of the inherent attention mechanism in GAT \cite{velickovic2017graph}, and the set of critical points in PointNet architectures \cite{qi2017pointnet}. The former, as originally proposed, however, produces attentions in between the nodes of the graphs, but does not offer explanations with regard to the prediction, and the latter is a simplification of the input point cloud, not indicative of the importance of the points to the model decision. In order to address this shortcoming, we propose the Class Node Graph Attention Network (CGAT), which can operate on graph representations of 3D shapes and can generate attention map-based explanations for model decisions. The CGAT architecture aims to provide competitive performance in graph-level tasks such as shape classification and regression, and we leverage the inherent learned attention of GAT convolution operation in order to depict the contribution of nodes to the final decision. To the best of our knowledge, this is the first study to propose an architecture capable of generating attention-based explanations for its predictions on 3D shapes. We demonstrate the CGAT architecture in the context of dental stage assessment of 3D third molar meshes, which is an important intermediate step in age estimation in forensics, in order to demonstrate the advantages of attention-based explanations in this high-stakes application. Dental stage assessment is the process of classifying individual teeth into developmental stages, which can then be utilized to predict the age of the subject \cite{rahim2023reliability}.  This process is conventionally carried out manually by experts assessing 2D radiographs. Herein lie two major drawbacks: inter-observer variability between the experts and a loss of information due to the projection of 3D structures onto a 2D image. 

To counter the first drawback, it is recommended that two observers assess the radiograph independently, and that they discuss in case of discrepancies to obtain a consensus \cite{de2020dental}. Thus, the main role of DL methods for dental staging lies in eliminating inter-observer variability. Although promising results have been published, no explainability mechanisms have been reported in the literature for this task \cite{matthijs2024artificial}, while plausibility of the staging is essential in a forensic context. After all, forensic age estimation implies a high-stakes decision where the misclassification of a minor as an adult represents an ethically unacceptable error. Therefore, explainability mechanisms of automated staging can contribute to a well-founded age estimation, helping the end user in understanding the outcome. To counter the second drawback, 3D imaging modalities have been studied for dental stage allocation \cite{franco2020comparing}. We chose to study CBCT because of its widespread use in dental practice, and thus an abundance of data \cite{brown2014basic}.  In the current study we aimed to design the explainability mechanism of CGAT to support the plausibility of automated dental stage allocation on CBCT-derived meshes. The CGAT architecture belongs to the GNN family, and while GNNs in general, including the CGAT, are not limited to graph-level tasks, and can also perform node-level and edge-level operations, for a broader shape-informed decision making framework, we limit the application of CGAT to graph-level classification in this study.

\section{Methodology}

\subsection{Data \& Preprocessing}

\begin{figure}[h!]
    \centering
    \includegraphics[width=\linewidth]{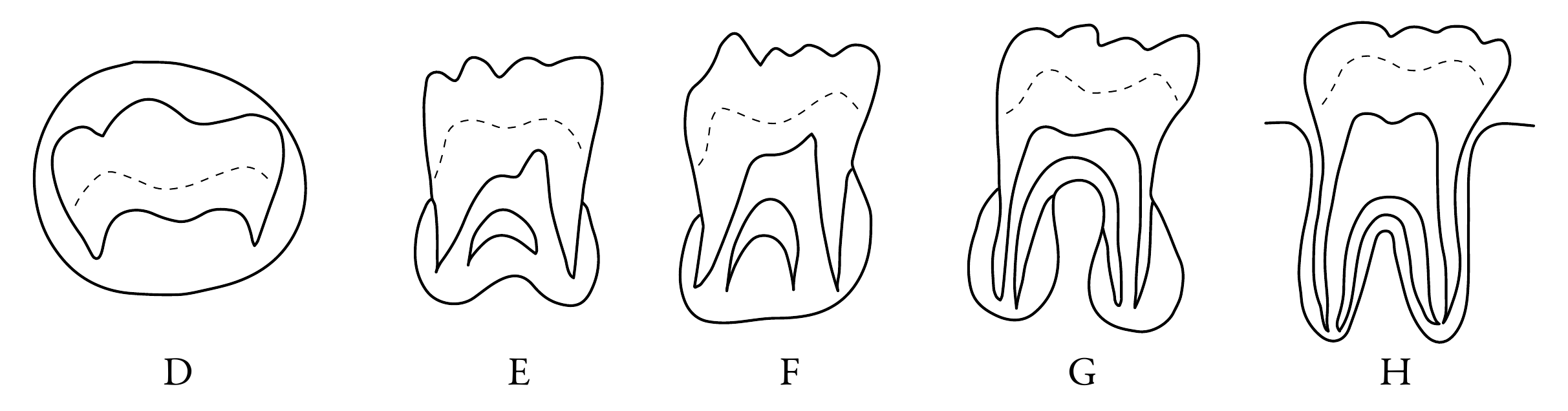}
    \caption{A visualization of the criteria for stage allocation of stages D to H according to the Demirjian assessment system \cite{demirjian1973new}. The staging criteria in this range depend mostly on root development. As the development progresses, the roots start to protrude from the crown and get longer, finally developing closed apices.}
    \label{fig:stageCrit}
\end{figure}

The dataset of focus for this study consists of triangular meshes of the four human third molars, corresponding to World Dental Federation (FDI) elements numbered 18, 28, 38 and 48 \cite{harris2005tooth}. The original data collection was performed at five Belgian centers by Vranckx et al. in the form of CBCT images of prospective patients for third molar surgery \cite{vranckx2021prophylactic}. Additional sample selection criteria were then applied on the 6010 patient records. This sample selection adhered to the criteria of patients aged 16 to 22 years old with at least two third molars, images containing no artifacts which affect third molars, such as metallic objects or presence of motion. Furthermore, third molars which were affected by pathological conditions were also excluded. The selection resulted in 138 patients in total, of which 67 were males and 71 were females, culminating in a total of 528 third molars. The third molars of the selected patients were then classified according to the developmental stage allocation system proposed by Demirjian et al. \cite{demirjian1973new}, by two dentists experienced in forensic dentistry and dentomaxillofacial radiology, respectively. The dentists carried out the stage assessment procedure on the corresponding panoramic radiographs for each CBCT recording \cite{franco2020comparing}. In the cases where the assessors did not agree on the stage, the dispute was resolved by an oral and maxillofacial radiologist with 8 years of experience. The third molars in the dataset ranged between Demirjian stages D to H, the criteria for which can be seen in Fig. \ref{fig:stageCrit}. 

\begin{figure}[!t]
    \centering
    \includegraphics[width=\linewidth]{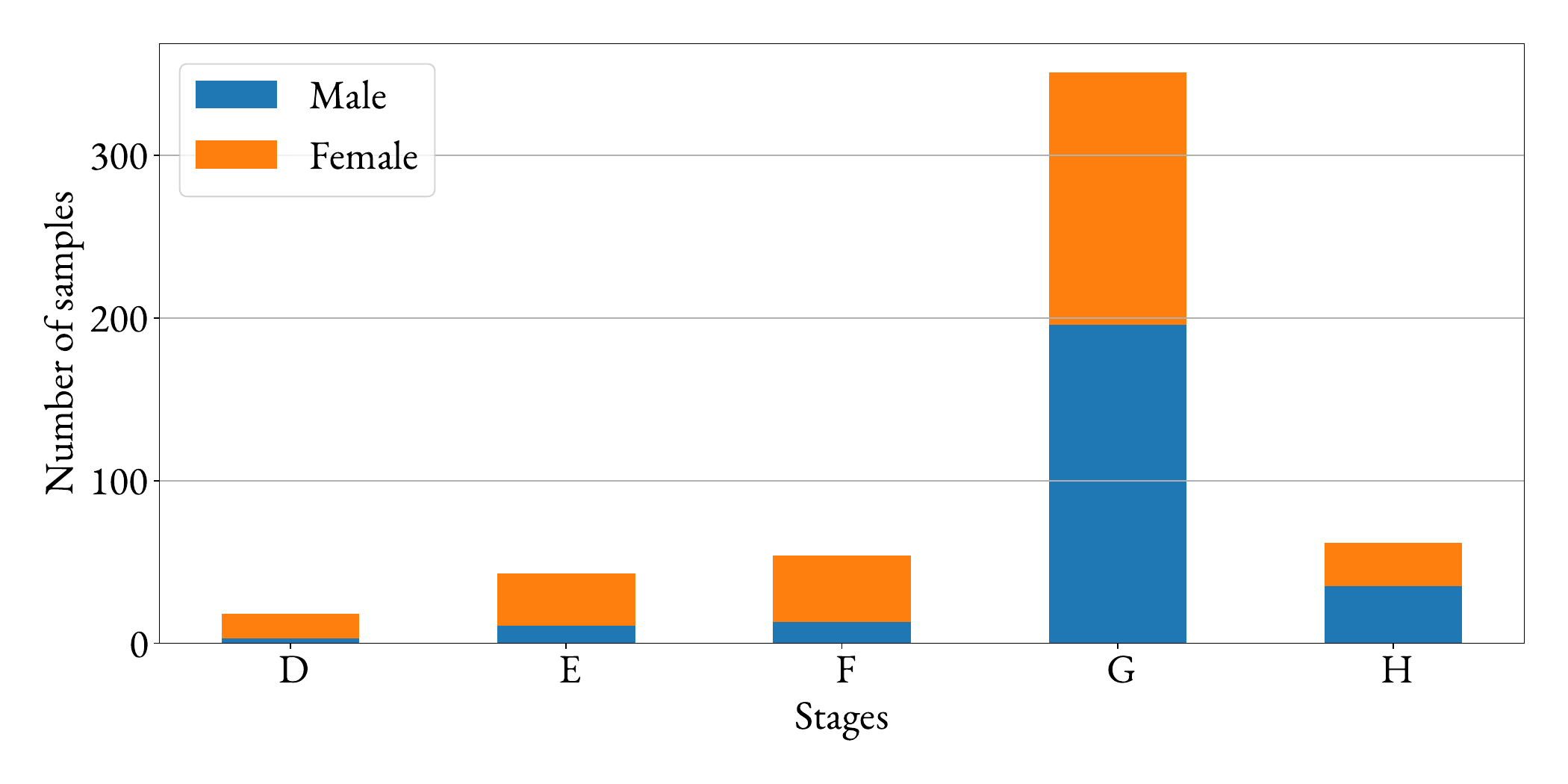}
    \caption{Distribution of samples over the Demirjian developmental stages. Note that the stages contain high class imbalance, with the lowest stage making up approximately 3\% of all data, and the stage containing most samples, G, accounting for 67\% of the dataset.}
    \label{fig:stage_distrib}
\end{figure}

These tooth samples contained high class imbalance, with the lowest representation belonging to the earliest stage, D, with only 18 samples, and the most well represented stage was G with 351 teeth. A distribution of samples over stages can be seen in Fig. \ref{fig:stage_distrib}. The CBCT images were then uploaded to the Relu\textsuperscript{\textregistered} Creator software \cite{reluCreator}, with which all third molars in the images were segmented, and the triangular meshes were created for each of them.

\begin{figure}
    \centering
    \includegraphics[width=\linewidth]{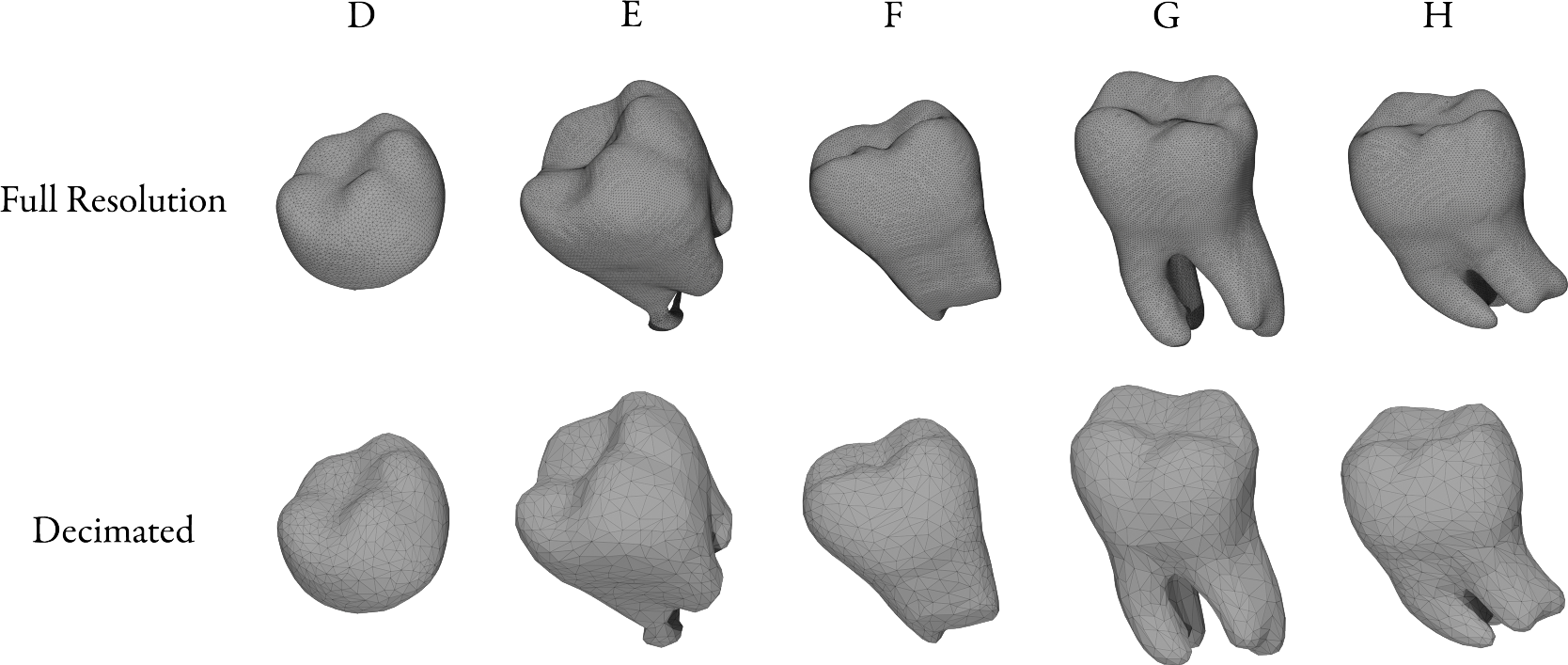}
    \caption{Visualizations of sample meshes from each Demirjian stage in full resolution (a), and their decimated counterparts (b). The decimation allows for the elimination of the negative effects of the substantial variation in the number of nodes and edges, and makes model training and computational analyses feasible. It can be seen that the original outputs of the Relu\textsuperscript{\textregistered} Creator software are overly dense, and the decimation step does not result in the loss of overall shape information.}
    \label{fig:meshes_fullDec}
\end{figure}

In this study, we define all meshes as undirected graphs, and as such each sample $G(\mathcal{V},\mathcal{E})$ is defined by a set of nodes $\mathcal{V} = \{v_1, v_2, ...,v_n\}$, and the edges $\mathcal{E} = \{e_1, e_2,...,e_m\}$ between these nodes, where $n$ and $m$ are the respective number of nodes and edges of each sample. The Relu\textsuperscript{\textregistered} Creator does not impose limits on the number of nodes and edges, and as such, the original outputs of the software had a varied number of these elements for each sample. More specifically, the output meshes consisted of $17205\pm5389$ nodes and $103225\pm32336$ edges on average. Thus, to remove the effect of this large variation in the number of graph elements, and to facilitate model training, these meshes were simplified using quadric mesh decimation to contain $751\pm1$ nodes, and $4497\pm4$ edges. Depictions of full resolution mesh samples from each category, along with their decimated counterparts can be seen in Fig. \ref{fig:meshes_fullDec}. It can be observed that the regions of interest for dental staging, e.g. structures such as the roots and the cusp are preserved in the decimated meshes, and shape information was not lost in these areas. 

In order to use the decimated meshes for the model training, all of them were then scaled such that they all fit into a unit sphere in Euclidean space. The scaling was motivated by the desire to train the model on the shape information alone. If left in the original relative sizes, the teeth from earlier stages of formation would be smaller than those with fully formed roots. This would introduce bias into the learning process and could result in the model exploiting this information, which we consider an undesirable effect for the accurate depiction of the attention maps, relating to the shape of the teeth. This decision is further supported by the intrinsic variation in the size of third molars; some early stage third molars may turn out rather large, while some late stage third molars may turn out rather small. The changes in shape, on the other hand, are more consistent across individuals. 

Upon preprocessing, 80\% of the samples were assigned to be used in training, 5\% was used in validation and 15\% was reserved for testing, preserving the distribution of stage labels.

\subsection{Network Architecture}
In order to facilitate the generation of attention maps on mesh classification, we employ the attention-weighted graph convolutions, first defined by Veličković et al. in \cite{velickovic2017graph}. The intrinsic attention mechanism allows for assigning weights to node features in graph convolutions through training. To generate attention-based explanations, we propose a modified architecture, which we refer to as the class-node graph attention network (CGAT). 

\subsubsection{Graph Attention Networks}
Graph attention networks were introduced to rectify a common shortcoming in other GNN architectures, which is the aggregation of the node features in local neighborhoods with uniform weights. More specifically, the general framework of GNNs is a two step process of applying a (usually learnable) transformation to each node feature $h_i, i \in \mathcal{V}$, and aggregating the features around the neighborhood $\mathcal{N}_i = \{j\in \mathcal{V} \mid e_{j\rightarrow i} \in \mathcal{E}\}$. One layer in a graph convolution operation then allows each node to collect information from its 1-hop neighborhood, and via repeated convolutions this receptive field expands. Many common GNN architectures use uniform-weighted aggregation operations such as averaging node features, or max-pooling. The GAT convolution operation introduces a weighted aggregation operation as

\begin{equation}\label{eq:gatlayer}
    {h_i}^{(l+1)} = \concat_{k=1}^K\sigma\left(\sum_{j : j \in \mathcal{N}_i}\alpha^{(l,k)}_{ij}.\boldsymbol{W}^{(l,k)}h_i^{(l)}\right)
\end{equation}

where $h^{(l,k)}$ is the node feature vector, $\alpha^{(l,k)}_{ij}$ is the learnable attention weight and $\boldsymbol{W}^{(l,k)}$ is the learnable transformation of attention head $k$ in layer $l$. The GAT thus defines a multi-head attention mechanism to increase the robustness of the learned attention \cite{velickovic2017graph,vaswani2017attention}. The learnable attention weights $\alpha^{(l,k)}_{ij}$ enable the higher consideration of some neighbors over others, and depend on the feature vectors of the neighboring nodes.

\begin{equation}
\label{eq:gatatt}
\begin{split}
    \alpha^{(l,k)}_{ij} &= \mathrm{softmax}(e(h_i,h_j))\\
    e(h_i,h_j) &= \text{LeakyReLU}\left(\boldsymbol{\alpha}_{(l,k)}^{\top} \cdot \left[\boldsymbol{W}^{(l,k)}h_i || \boldsymbol{W}^{(l,k)}h_j \right]\right)    
\end{split}
\end{equation}

In Eq. \ref{eq:gatatt}, the $\boldsymbol{\alpha}_{(l,k)}$ is the learnable attention parameter of layer $l$, from attention head $k$. However, this attention mechanism has diminished expressiveness, due to it being a form of static attention as defined by Brody et al. \cite{brody2021attentive}, where regardless of the target node, there is always a source node that will have the highest contribution in the aggregation operation. They propose a modification, dubbed GATv2, that reorders the application of the transformation $\boldsymbol{W}^{(l,k)}$ and the attention parameter $\boldsymbol{\alpha}_{(l,k)}$ as,

\begin{equation}\label{eq:gatv2conv}
    e(h_i,h_j) = \boldsymbol{\alpha}_{(l,k)}^\top \text{LeakyReLU}\left( \boldsymbol{W}^{(l,k)}\cdot\left[h_i||h_j\right]\right)
\end{equation}

therefore rectifying the consecutive application of $\boldsymbol{W}^{(l,k)}$ and $\boldsymbol{\alpha}_{(l,k)}$, and using the attention parameter as an independent transform. In our experiments, we adopt the GATv2 convolution operation due to the improvements over the original proposal from \cite{velickovic2017graph} shown in Eq. \ref{eq:gatatt}.

\subsubsection{Class Node Graph Attention Network}

\begin{figure}[H]
    \centering
    \includegraphics[width=\linewidth]{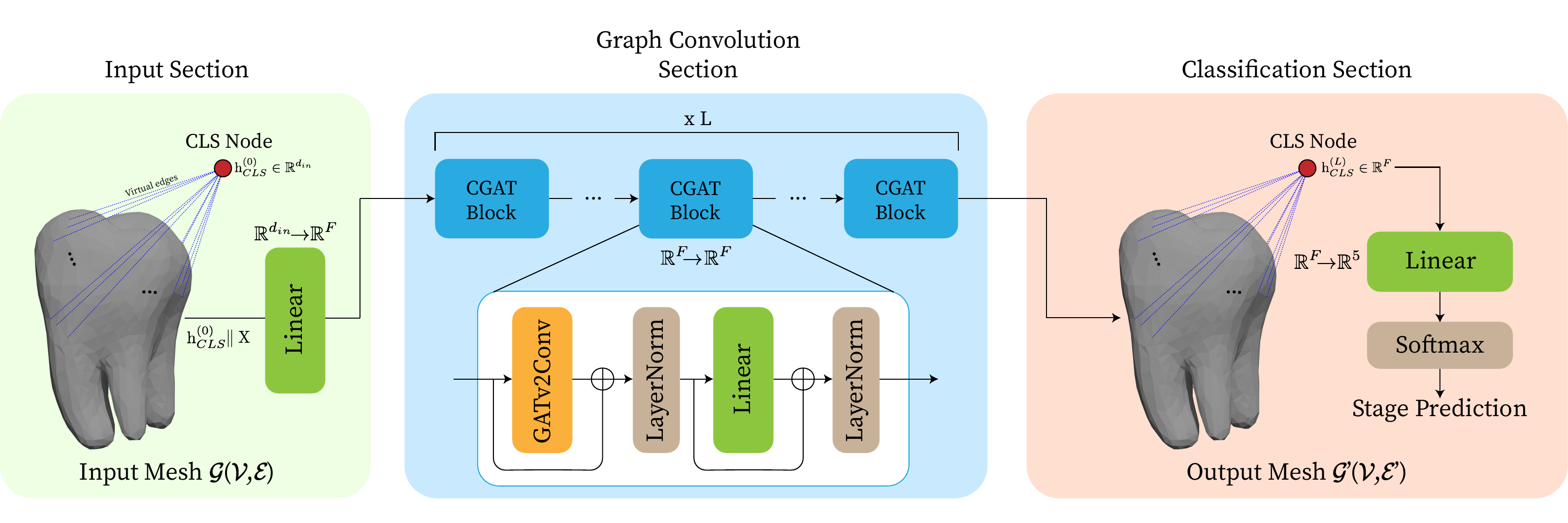}
    \caption{Overview of the CGAT architecture. The overall architecture consists of three sections: (1) The input section is responsible for connecting the virtual CLS node to all nodes in the input graph, and applying a linear projection of all node features. (2) The graph convolution section of the model consists of stacked CGAT blocks, and is responsible for performing the attention-informed graph convolutions. (3) The classification section extracts the final embedding of the CLS node and applies one linear layer and a softmax function to produce the probability predictions of the stages. The architecture uses the attention scores of the CLS node to all other nodes to produce attention-based explanations.}
    \label{fig:cgat}
\end{figure}

We propose the CGAT architecture in order to introduce explainability in the classification of the tooth meshes, in the form of attention maps. We leverage the intrinsic attention mechanism in the GATv2 convolution in order to do so, by the introduction of a virtual node, which we call the \textit{CLS node} throughout this paper. The addition of a virtual node as a global information aggregator is an existing design philosophy. Among transformer models, the [CLS] token in BERT \cite{devlin2019bert}, Vision Transformer (ViT) \cite{dosovitskiy2020image} and the Graphormer \cite{ying2021transformers} are notable examples implementing this design, the latter of which inspired by the usage of a \textit{master node} by Gilmer et al. \cite{gilmer2017neural}. To the best of our knowledge, however, ours is the first study to use such a master node for the generation of attention maps on 3D shapes. 

The core mechanic of the CGAT is the introduction of this virtual node as the first step of training to each input graph, individually. The CLS node is connected to each node in the graph, and treated no differently from the other nodes in the input graph throughout the graph-convolutional section of our model. The initial embedding of the CLS node $h^0_{\text{CLS}}$ is a learnable parameter of the model. We experiment with connecting the CLS node to input graphs with directed edges $e_{i:i\in\mathcal{V} \rightarrow \text{CLS}}$ which flow only towards the CLS node, and undirected edges $e_{i:i\in\mathcal{V} \leftrightarrow \text{CLS}}$. This selection dictates whether the learned latent representations of the CLS node throughout the model can influence the latent embeddings of each node in the graph. We report on the effects of this choice in Section \ref{sec:results}. 

The CGAT consists of three functional sections; (1) the input section, where the input graphs are accepted, and the CLS node is appended to each of them, (2) the graph convolution section, where the CGAT blocks reside, and (3) the classification section, where the CLS node embedding is extracted and fed to a multilayer perceptron (MLP) head to reach the final classification output. An overview of this architecture can be seen in Fig. \ref{fig:cgat}. 

The input section of the model is only responsible for projecting all node features onto a latent space, and correctly appending one virtual node to each input graph in the batch. We note that regardless of the choice of using directed or undirected edges to connect the CLS node to the graph, all nodes \textit{except for the CLS node} are also assigned a self-connection in order to ensure that the self-attention does not overshadow the attention to all other nodes, since the input degree of the CLS node $D^{\text{CLS}}_{in} >> D_{in}^{v_i\in\mathcal{V}}$ is much larger than the average input degree of the original graph.
We batch the input graphs, where each graph corresponds to a single tooth, by merging each sample to form an unconnected graph with a block-diagonal adjacency matrix. The input projection consists of one linear layer parametrized by $\boldsymbol{W}_{in}\in\mathbb{R}^{d_{in}\times F}$ where $d$ is the unprojected dimensions of the node features $x \in \mathbb{R}^{d_{in}}$. The CLS node embedding $h^0_{CLS}\in\mathbb{R}^{F}$, as mentioned, is a learnable parameter of the model. We used the Glorot initialization for this parameter \cite{glorot2010understanding}, as with all learnable parameters of the CGAT model. We allow the learning of $h^0_{\text{CLS}}$ in order to find the best starting position in the input latent space via backpropagation.

\begin{algorithm}[h!]
\caption{CGAT Forward Pass}
\label{alg:cgat_forward}
\begin{algorithmic}[1]
\Require Graph $G(\mathcal{V}, \mathcal{E})$, node features $X \in \mathbb{R}^{|\mathcal{V}| \times d_{in}}$
\Ensure Class logits $\mathbf{y}_{\text{logits}} \in \mathbb{R}^{N_{\text{classes}}}$

\State \Comment{--- Preprocessing \& Input Section ---}
\State Augment $\mathcal{E}$ with self-loops for all $v \in \mathcal{V}$
\State $N \gets |\mathcal{V}|$
\State Augment $\mathcal{V}$ with a new CLS node: $\mathcal{V}' \gets \mathcal{V} \cup \{v_{\text{CLS}}\}$
\State Augment $\mathcal{E}$ to connect all nodes in $\mathcal{V}$ to $v_{\text{CLS}}$: $\mathcal{E}' \gets \mathcal{E} \cup \{e_{i {(\leftrightarrow ,\rightarrow)} \text{CLS}}\}_{i=1}^{N}$
\State Initialize learnable CLS embedding: $\mathbf{h}_{\text{CLS}}^{(0)} \gets \mathbf{\Theta}_{\text{CLS}} \in \mathbb{R}^{1 \times d_{in}}$
\State Concatenate node features with CLS embedding: $X_{\text{aug}} \gets \text{Concat}(X, \mathbf{h}_{\text{CLS}}^{(0)}) \in \mathbb{R}^{(N+1) \times d_{in}}$
\State Project features into latent space: $\mathbf{H}^{(0)} \gets \text{Linear}(X_{\text{aug}}) \in \mathbb{R}^{(N+1) \times F}$

\State \Comment{--- Graph Convolution Section ---}
\For{$l = 0$ to $L-1$}
    \State $\mathbf{H}_{\text{conv}} \gets \text{GATv2Conv}(\mathbf{H}^{(l)}, \mathcal{E}')$ \Comment{Multi-head attention as in Eq. 4}
    \State $\mathbf{H}_{\text{res1}} \gets \text{LayerNorm}(\mathbf{H}^{(l)} + \mathbf{H}_{\text{conv}})$ \Comment{First residual connection}
    \State $\mathbf{H}_{\text{lin}} \gets \text{Linear}(\mathbf{H}_{\text{res1}})$
    \State $\mathbf{H}^{(l+1)} \gets \text{LayerNorm}(\mathbf{H}_{\text{res1}} + \mathbf{H}_{\text{lin}})$ \Comment{Second residual connection}
\EndFor

\State \Comment{--- Classification Section ---}
\State Extract final CLS embedding: $\mathbf{h}_{\text{CLS}}^{(L)} \gets \mathbf{H}^{(L)}[N+1, :]$
\State Apply dropout: $\mathbf{h}_{\text{CLS}}^{(L)} \gets \text{Dropout}(\mathbf{h}_{\text{CLS}}^{(L)})$
\State Compute logits: $\mathbf{y}_{\text{logits}} \gets \text{MLP}(\mathbf{h}_{\text{CLS}}^{(L)})$ \Comment{Classification head}

\State \Return $\mathbf{y}_{\text{logits}}$
\end{algorithmic}
\end{algorithm}



The batched inputs with the CLS node appended and their node features projected are then fed into the graph convolution section of the CGAT. This section consists of sequentially stacked CGAT blocks, where each block is made up of one GATv2 convolution layer, and one linear layer (Fig. \ref{fig:cgat}). It is worth noting that the term \textit{block} for the CGAT architecture corresponds to the term \textit{layer} in traditional GNN models. Similar to the GAT architecture, we employ multi-head attention in the GATv2 convolutions throughout the graph convolution section with $K=8$ attention heads. Unlike the GAT architecture, however, where individual attention head outputs are concatenated, or averaged for the final layer, we merge attention heads with max-pooling, where only the maximum value of each node feature is kept.

\begin{equation}
\label{eq:cgatblock}
\begin{gathered}
    h_i^{(l+1)} = \text{LayerNorm}\left(
    {h'}_i^{(l)} + \sigma_2\boldsymbol{W}_2^{(l)}\cdot {h'}^{(l)}_i
    \right)\\
    {h'}^{(l)}_i = \text{max}(k_{1:K})\left[\text{LayerNorm}\left(
    h_i^{(l)} + \sigma_1\sum_{j:j\in\mathcal{N}_i} \alpha_{ij}^{(l,k)}.\boldsymbol{W}_1^{(l,k)}h_i^{(l)}
    \right)
    \right]
\end{gathered}
\end{equation}

Eq. \ref{eq:cgatblock} describes a single CGAT block, where $\boldsymbol{W}^{(l,k)}_1\in\mathbb{R}^{F\times F}$ is the shared weight matrix from Eq. \ref{eq:gatv2conv} of head $k$ of block $l$, $\boldsymbol{W}^{(l)}_2\in\mathbb{R}^{F\times F}$ is the linear transform applied after the GATv2 convolution of block $l$, ${h'}^{(l)}_i$ is the output of the graph convolution, and $\sigma_1, \sigma_2$ are the nonlinearity functions, selected as the GeLU nonlinearity by default. We include residual connections around the graph convolutions and linear layers in order to rectify the vanishing gradient problem. The structure of the CGAT block is inspired by the transformer architecture from \cite{vaswani2017attention}, and is designed to increase the expressive capability of the CGAT model by introducing additional parameters. In this sense, a CGAT block can be viewed as a transformer layer, using dynamic attention from \cite{brody2021attentive} instead of the scaled dot product attention used by Vaswani et al., and without the positional encoding step. The graph convolution section of the CGAT architecture is then built by sequential application of block $l\in L$ to the output of the previous section, as detailed in lines 9-13 in Algorithm \ref{alg:cgat_forward}.

The classification section of the CGAT model, which follows the graph convolution section, takes the learned CLS node embedding $h^{(l+1)}_{\text{CLS}}$ after $l$ blocks as \textit{the only} input, therefore ensuring whatever information the final decision is based on flows through the CLS node, thus making the CLS node a global information aggregator. We employ an MLP head for classification parametrized by $\boldsymbol{W}_{\text{MLP}} \in \mathbb{R}^{F\times 5}$, mapping $h^{(l+1)}_{\text{CLS}}$ to class logits to be used in the loss calculation. We further apply the softmax function to the logits to get class probabilities. 

\begin{equation}
    \tilde{y} = \mathrm{softmax}(h^{(l+1)}_{\text{CLS}}\cdot\boldsymbol{W}_{\text{MLP}})
\end{equation}

Throughout the three sections described, the model learns to produce CLS node embeddings that are informative of the developmental stage of the teeth, based on the input node features and the graph connectivity. The general aim of the CGAT architecture is to produce human-understandable representations of attention while preserving the predictive power of the GAT family of GNNs. In order to generate such representations, we leverage the attention weights. 

\subsubsection{Attention Map Generation}
We design CGAT so that the information is funneled through the CLS node to the classification section. This design decision is the key factor in the ability of the CGAT to produce attention maps, where the learned attention scores with the target node as $v_{\text{CLS}}$ can be utilized as a means of analyzing the contribution of each node to the model decision. Thanks to the inherent attention mechanism of the GATv2 convolutions, the visualization of attention maps to represent said contribution becomes possible. In order to bring about visual depictions of the regions of attention, we employ the attention rollout technique \cite{abnar2020quantifying}. The attention rollout method was first proposed to display the attention of transformers with many layers, and has also been employed to depict the learned attention of the ViT models on 2D images in \cite{dosovitskiy2020image}. 
The attention rollout method can be formulated as,
\begin{equation}
\begin{split}
    \tilde{\boldsymbol{A}}^{(l)} &= (\boldsymbol{A}^{(l)}+\boldsymbol{I})\tilde{\boldsymbol{A}}^{(l-1)}\\
    \tilde{\boldsymbol{A}}^{(0)} &= \boldsymbol{A}^{(0)} + \boldsymbol{I}
\end{split}
\end{equation}

where $\tilde{\boldsymbol{A}}^{(l)}$ is the rollout and $\boldsymbol{A}^{(l)}\in\mathbb{R}^{n+1\times n+1}$ is a sparse matrix containing the learned attention scores corresponding to each edge in the graph, of layer $l$, and $\boldsymbol{I}$ is the identity matrix, modeling the residual connections. With this recursive operation, the "flow" of the attention from the CLS node to each input node can be represented holistically by visualizing only the values $a_{\text{CLS}\rightarrow i} : e_{i\rightarrow\text{CLS}}\in\tilde{\mathcal{E}}$ on the corresponding nodes in the input meshes. Since our application is a graph-level classification task, we do not analyze the attentions between the nodes of the input graph $a_{j\rightarrow i}:e_{i\rightarrow j} \in {\mathcal{E}\cap\tilde{\mathcal{E}}}$. These values are only used in the calculation of the attention rollout. 

In order to visualize the attention in an understandable manner, we performed an upper value clipping on the attention maps, where the values are clipped to the range of $[min(\tilde{\boldsymbol{A}}),mean(\tilde{\boldsymbol{A}}) + min(\tilde{\boldsymbol{A}})]$. To determine these limits, we use the concatenated rollout scores from all the samples visualized in order to preserve the relative scales.

\subsubsection{Model Training and Evaluation} 
In order to rectify the effects of large class imbalance, we employed weighted sampling in batch creation so that each batch represented a class-wise balanced selection of samples. The training optimized for the cross-entropy loss. The Adam optimizer was selected for all trainings with an initial learning rate of 0.001. A learning rate scheduler with a patience of 5 epochs was used to reduce the learning rate by a factor of 0.5 if no improvement was observed in the validation loss within the patience range. All models were trained for 150 epochs. A dropout operation with $p=0.3$ probability was applied on the CLS node embedding after the graph convolutional section in order to increase the robustness of training against overfitting. 

For the evaluation of the model performance, following a multiclass classification, the weighted F1 score was selected. This metric is an effective descriptor of classification capabilities in the presence of high class imbalance. The weighted F1 score is defined as,

\begin{equation}
    \begin{split}
    \text{Weighted F1} = \sum_C w_c\text{F1}_c &= \sum_C w_c \left(\frac{\text{TP}_c}{\text{TP}_c + 1/2(\text{FP}_c + \text{FN}_c)}\right)\\
    w_c &= \frac{\|c\|}{\|C\|}      
    \end{split}
\end{equation}

where $\text{TP}_c$, $\text{FN}_c$ and $\text{FP}_c$ stand for one-versus-rest true positives, false negatives and false positives, $w_c$ signifies the class weight, or ratio of the number of samples to the total number of samples for class $c$. The weighted F1 score is a useful metric which summarizes the precision and recall values, with regard to class imbalance. In addition to the metric evaluation, we perform qualitative evaluations of the attention maps generated by the model. We regard the conformity of these attention maps to human understanding to be of higher priority than the classification metrics, in the context of this paper. Although the metric performance of the model is important, if the models do not show attention patterns that overlap with expert knowledge to some extent, then the question of trust in deep models remains unanswered.

As for the software and hardware used in the implementation of CGAT and the experiments, all training and evaluation was done using the PyTorch and PyTorch Geometric frameworks with GPU accelerated computing with NVIDIA CUDA, and an NVIDIA A100 GPU with 80GB of memory was utilized for computations. 

\subsection{Node Features}
The selection of node features, while important for the classification performance, is more so for the visualization of attention, and the generation of human-understandable representation of the attention. Since the only available information to CGAT are the node features $X = \{x_1,...x_n,...,x_N\}$, and the presence of edges, the final attentions are learned \textit{only} with respect to these elements. In order to approximate the human understanding of 3-dimensional shapes in the model attention, the selected node features must also reflect the qualities of the shapes that an expert would focus on. In dental stage allocation within the given range of stages in our dataset (D to H), the focus of experts is on the development of the root(s), and the closing of the root apices (Fig. \ref{fig:stageCrit}). In order to represent the shape information of these areas in the local neighborhood, we encode each node in the input graphs with shape-relevant features, namely the mean curvature and the distance to mesh centroid. These features are selected due to their pose-invariant quality, since the samples in the dataset are not in alignment, pose-dependent node features such as coordinates would lead models to depend on the position of the meshes in space, and therefore disregard shape information. Due to the alignment of varied shapes such as teeth being non-trivial, we favor rotation and translation-invariant learning via the employment of these features instead of performing mesh alignment. Furthermore, after the calculation of all features we apply min-max scaling across all samples so that all features are in the $[-1,1]$ range while preserving the relative differences between the samples.

\subsubsection{Mean Curvature}
\begin{figure}[!htp]
    \centering
    \includegraphics[width=\linewidth]{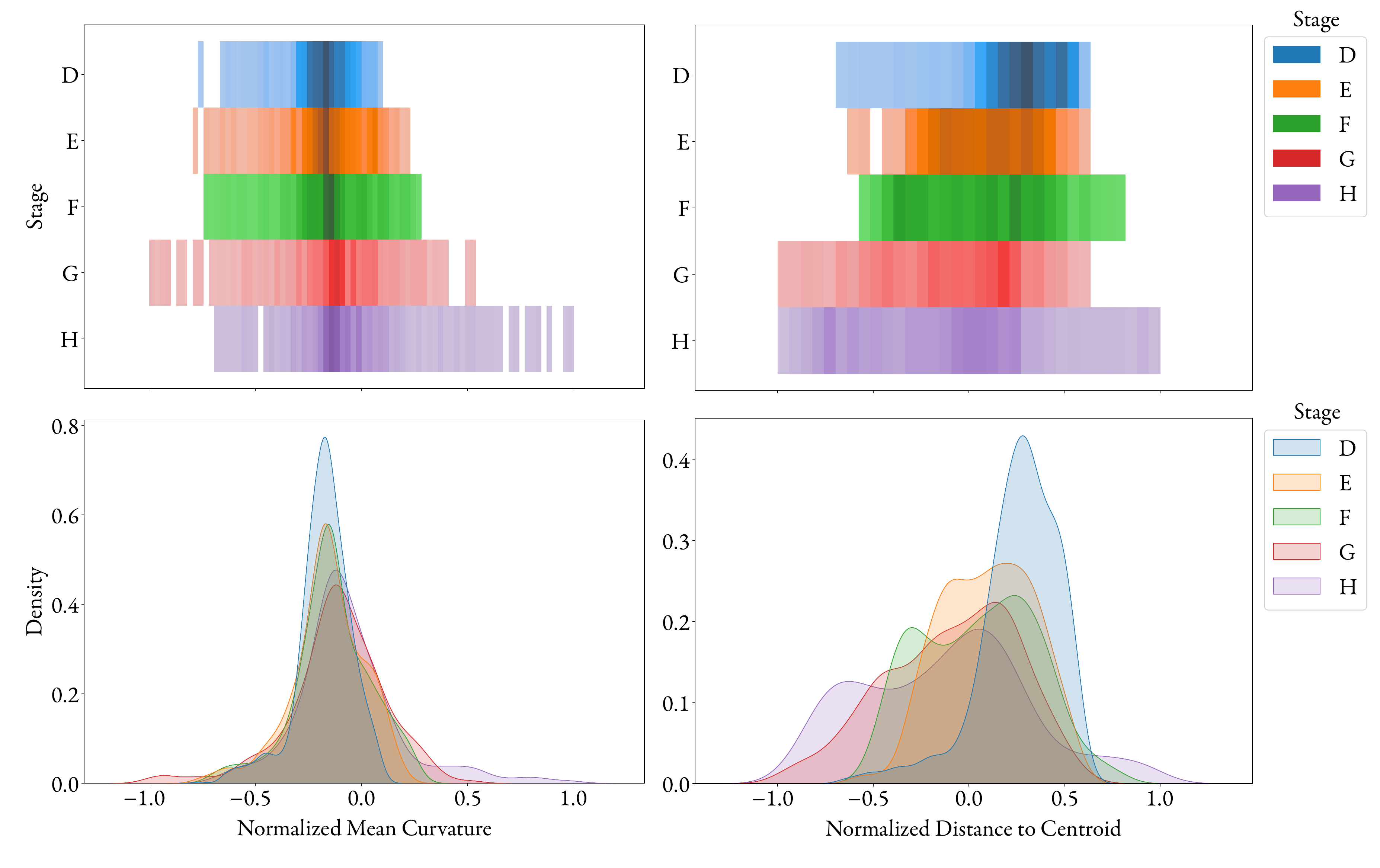}
    \caption{Distributions of the normalized mean curvature feature for all nodes, separated by Demirjian stage. As dental development progresses from stage D to H, the distribution skews, showing an increase in nodes with high positive curvature. This corresponds to the formation of convex shapes like root apices.  The clear separation of these distributions demonstrates that mean curvature is a highly discriminative feature, providing a strong signal for the classification task which can be leveraged by the CGAT.}
    \label{fig:featDist}
\end{figure}

The mean curvature of a point on a surface is a function of the principal curvatures $k_1$ and $k_2$, and when calculated locally around each node is indicative of the concavity of the neighborhood. Mean curvature, thus, is a highly desirable descriptor for teeth, as the third molars tend to have less peaks due to the lack of the root apices in the early development, and a high mean curvature value in a node can be indicative of the development of the roots. Moreover, as the teeth get more elongated through subsequent stages, they contain more flat surfaces with very low curvedness, which again can be captured by the mean curvature feature. The difference in the distributions of the mean curvature feature encoded in the nodes of each label is shown in Fig. \ref{fig:featDist}. These distributions show that the features change significantly for different stages, and display the necessary discriminative qualities.

\subsubsection{Distance to Centroid}
The Euclidean distances of each node to the mesh centroid can be informative towards the overall roundness and elongation of the meshes. In the earlier stages, the teeth are more rounded due to the lack of roots, and as roots form during dental development, the nodes in the root area get further away from the centroid. As the centroid shifts due to the elongation throughout the stages, the distances of the nodes remain informative of their relative positions in the mesh.  We calculate this distance on the meshes after the preprocessing is carried out, with all meshes simplified and scaled into a unit sphere. Fig. \ref{fig:featDist} shows the distributions of the normalized features from all nodes per class, which, similarly to the distributions of the mean curvature, differ across the developmental stages.

\section{Experiments and Results}\label{sec:results}
\renewcommand{\arraystretch}{1.5}
\begin{table}[!hptb]
\caption{The metric results of the depth evaluation of CGAT variants. For each model family signifier, weighted F1 score is reported in the following row, where each columnar entry signifies the value of the metric corresponding to the model with the specified number of blocks. We evaluate models with 1 to 15 blocks, trained and tested using the mean curvature and the distance to centroid as node features, and another family of models using both these features. Each model is trained 10 times and the average value is reported. The standard deviations are not reported for the sake of brevity. The maximum value in each row, and values within the $\pm$ 0.01 range of it, are shown in bold for emphasis. }\label{tab:results}
\scriptsize
\centering
\begin{tabular}{|c|c|c|c|c|c|c|c|c|c|c|c|c|c|c|c|}
\hline            
\diagbox[height = 3\line,innerwidth=2.3cm ]{\tiny{\textbf{Model Family}}}{\tiny{\textbf{N.Layers}}}                             & \textbf{1} & \textbf{2} & \textbf{3} & \textbf{4} & \textbf{5} & \textbf{6} & \textbf{7} & \textbf{8} & \textbf{9} & \textbf{10} & \textbf{11} & \textbf{12} & \textbf{13} & \textbf{14} & \textbf{15} \\ \hline

{${}^{1:15}\text{CGAT}^{\leftrightarrow}_{curv}$} & 0.58 & 0.61 & 0.58 & 0.65 & 0.68 & 0.63 & 0.68 & 0.68 & 0.66 & 0.69 & 0.67 & 0.68 & \textbf{0.71} & 0.68 & \textbf{0.72}     \\ \hline

{${}^{1:15}\text{CGAT}^{\leftrightarrow}_{dist}$} & 0.66 & \textbf{0.70} & \textbf{0.71} & 0.66 & 0.68 & 0.64 & 0.66 & 0.67 & \textbf{0.70} & 0.67 & \textbf{0.70} & 0.66 & 0.69 & 0.66 & 0.68     \\ \hline

{${}^{1:15}\text{CGAT}^{\leftrightarrow}_{both}$} & 0.68 & 0.69 & 0.68 & 0.69 & 0.70 & 0.71 & 0.73 & 0.71 & 0.74 & \textbf{0.76} & 0.70 & \textbf{0.76} & 0.71 & 0.70 & 0.68    \\ \hline

{${}^{1:15}\text{CGAT}^{\rightarrow}_{curv}$} & 0.62 & 0.60 & 0.62 & 0.62 & 0.66 & 0.68 & 0.65 & 0.72 & 0.69 & 0.69 &\textbf{0.75} & 0.71 & 0.64 & 0.60 & 0.66    \\ \hline
{${}^{1:15}\text{CGAT}^{\rightarrow}_{dist}$} & 0.67 & \textbf{0.68} & \textbf{0.69} & \textbf{0.68} & 0.67 & \textbf{0.69} & 0.67 & 0.66 & 0.67 & 0.63 & 0.66 & \textbf{0.69} & \textbf{0.69} & \textbf{0.68} & 0.65      \\ \hline

{${}^{1:15}\text{CGAT}^{\rightarrow}_{both}$} & 0.66 & 0.66 & 0.67 & 0.70 & 0.69 & \textbf{0.76} & 0.70 & 0.73 & 0.70 & 0.73 & 0.72 & 0.70 & 0.68 & 0.71 & 0.68     \\ \hline                                                   
\end{tabular}
\end{table}

We devised several experiments in order to evaluate the attention maps and the predictive capability of the CGAT models, and demonstrate the effects of model depth and the direction of CLS edges. Throughout the rest of this paper, we refer to the models with the shorthand notation ${}^{L}\text{CGAT}^{E}_{F}$ for ease of discussion, where we denote the number of blocks $L\in\{1,2,\cdots,15\}$, the node features used $F\in\{curv, dist, both\}$ where \textit{curv} stands for mean curvature,  \textit{dist} stands for the distance to centroid and \textit{both} stands for a combination of both features and the type of the added edges that connect the CLS node to all edges $E\in\{\rightarrow,\leftrightarrow\}$ where $\rightarrow$ signifies directed edges from each node to the CLS node, and $\leftrightarrow$ indicates undirected edges, using superscripts and subscripts. As an example, ${}^{3}\text{CGAT}^{\rightarrow}_{curv}$ implies a model with 3 CGAT blocks, using mean curvatures as node features, connecting the CLS node to each node with directed edges $e_{i\rightarrow\text{CLS}}$.

\begin{figure}[!ht]
    \centering
    \includegraphics[width=\linewidth]{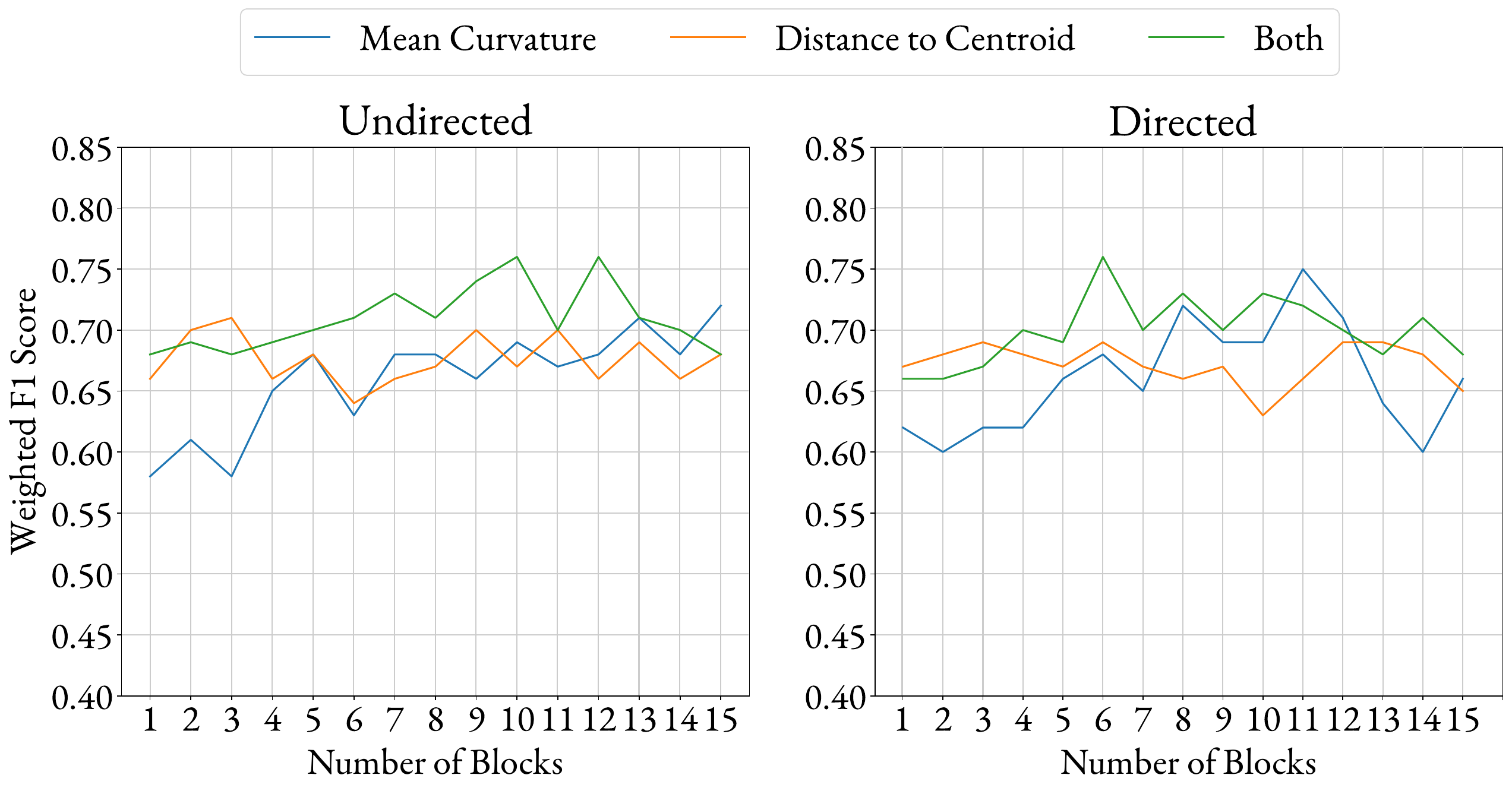}
    \caption{Plots of the weighted F1 scores of the models with the number of blocks ranging from 1 to 15, trained using the mean curvature, the distance to centroid and both features combined. Overall, the CGAT model performance does not deteriorate significantly in the given range of blocks, and the effects of over-smoothing are not observed. There is a slight upward trend observed in the metrics with undirected edges, and with directed edges the performance metrics are mostly stable across differing number of blocks. }
    \label{fig:metrics}
\end{figure}

For the experiments concerning the model depth and CLS edge direction, we compare the effects of the selected node features by training the models on mean curvature and the distance to centroid individually, and we also concatenate these features and train using the combined feature vectors. All the models, regardless of the other hyperparameters, have $K=8$ attention heads. Fig. \ref{fig:metrics} shows the weighted F1 score metrics for models with 1 to 15 blocks to demonstrate the effects of model depth. All models are trained using max-pooling to merge attention heads. Furthermore, we train all models with directed and undirected edges to the CLS node, and report on the resulting performance metrics. These results, displayed in Table \ref{tab:results} are discussed and the learned attention weights are depicted in the form of attention maps in the remainder of this section.

\subsection{Model Depth and Edge Direction}
GNNs are affected by the over-smoothing issue, where over multiple layers, the node features become more and more similar to each other due to continuous neighborhood aggregations. As a result, deep GNNs usually show a decaying performance as the number of layers increases \cite{rusch2023survey}. It is therefore important to evaluate a number of CGAT models with varying number of layers. We train and compare the performances of models with the number of blocks increasing from 1 to 15 with unit step. We then demonstrate and discuss the effect of model depth on the generated attention maps, using a subset of this model pool to visualize learned attentions. Fig. \ref{fig:metrics} shows the weighted F1 score metrics over the number of blocks. 
The direction of the edges used to connect the CLS node to all the other nodes in the input graphs has a direct impact on the learning process. If the edges are directed (from each node to the CLS node), the CLS node is allowed to attend to each node, and is only allowed to accumulate node features by the learned attention weights. Therefore, the embedding of the CLS node does not influence the node embeddings through aggregation.  In this setting, the CLS node has an in-degree of $d_{\text{CLS}}^{in} = n$ and an out-degree of $d_{\text{CLS}}^{out} = 0$. However, if undirected edges are used, the learned embedding of the CLS node is also aggregated into each node along with the neighbors of the node, therefore, the node embeddings $h_i^{(l)}$ and $h_{\text{CLS}}^{(l)}$ both contribute to each other to learn an embedding that represents the stage information. Since the nodes in our input graphs have, on average, a degree of $\bar{d}_i^{in} = \bar{d}_i^{out} = 7$, the contribution of the CLS node in the aggregation step is significant compared to the contribution of the node embeddings to the CLS node, since $\bar{d}_i^{in}<<\bar{d}^{in}_{\text{CLS}} = 751 $. Furthermore, by using undirected CLS edges, all nodes in the graph are made to be in the 2-hop neighborhood of each other. This connectivity scheme, when 2 or more blocks are implemented, allows the information of each node to influence all other nodes, regardless of mesh topology and graph distance. Due to these fundamental differences brought on by the direction of the edges, it is important to analyze the effects on predictive performance and the attention maps.

It can be seen from the results that the CGAT shows a slight increase in performance as the model depth is increased from 1 to 15 when undirected CLS edges $e_{i\leftrightarrow\text{CLS}}$ are used. With the ${}^{1:15}\text{CGAT}_{curv}^{\leftrightarrow}$ family of models, the best mean weighted F1 scores are encountered with 13 and 15 blocks. A similar case is also seen with the models trained with both features, where ${}^{10,12}\text{CGAT}_{both}^{\leftrightarrow}$ show the best weighted F1 metrics, and F1 scores above 0.7 are seen with block numbers 4 through 14. As for the  models with undirected edges using the distance to centroid as node features, a uniform performance profile can be observed where the variation of the mean metrics remains low across the number of blocks. These results indicate that the models with undirected edges did not suffer from performance loss due to oversmoothing in the given range of blocks. Contrastingly, ${}^{1:15}\text{CGAT}_{curv}^{\leftrightarrow}$ family of models consistently show a worse performance with ${}^{1:3}\text{CGAT}_{curv}^{\leftrightarrow}$ displaying the worst classification metrics, and are seen to benefit from an increased number of blocks, resulting in a larger receptive field for each node. Given all these observations, the best operating range for models with undirected edges can be said to be with 9 blocks and more.  

When directed edges $e_{i\rightarrow\text{CLS}}$ are used to connect the CLS edges, a roughly stable trend in the metrics can be observed as the number of blocks is increased. The CGAT models trained with the mean curvature deviate from this observation the most, as the models ${}^{1:11}\text{CGAT}_{curv}^{\rightarrow}$ seem to benefit from the increased depth, but the performance starts deteriorating with more depth, with ${}^{14}\text{CGAT}_{curv}^{\rightarrow}$ showing the same mean weighted accuracy as ${}^{2}\text{CGAT}_{curv}^{\rightarrow}$, hinting at the initial symptoms of oversmoothing. The model family which uses both features, ${}^{1:15}\text{CGAT}_{both}^{\rightarrow}$ also mimics this trend, albeit less severely. The metrics increase for these models up until 6 blocks, after which a plateau in performance can be observed. As for the models trained with distance to centroid, we see a monotonic performance profile, with models yielding mean weighted F1 scores that are all in the $[0.63,0.69]$ range, with the highest performances reported with shallower models as well as deeper ones. As per the results of the depth experiments, we suggest an operating range of 6 to 12 blocks, regardless of the node feature used. For both types of CLS edges, using both features together, instead of individually, shows better performance in the given operating ranges.

\subsection{Attention Maps}
\begin{figure}[h]
    \centering
    \includegraphics[width=\linewidth]{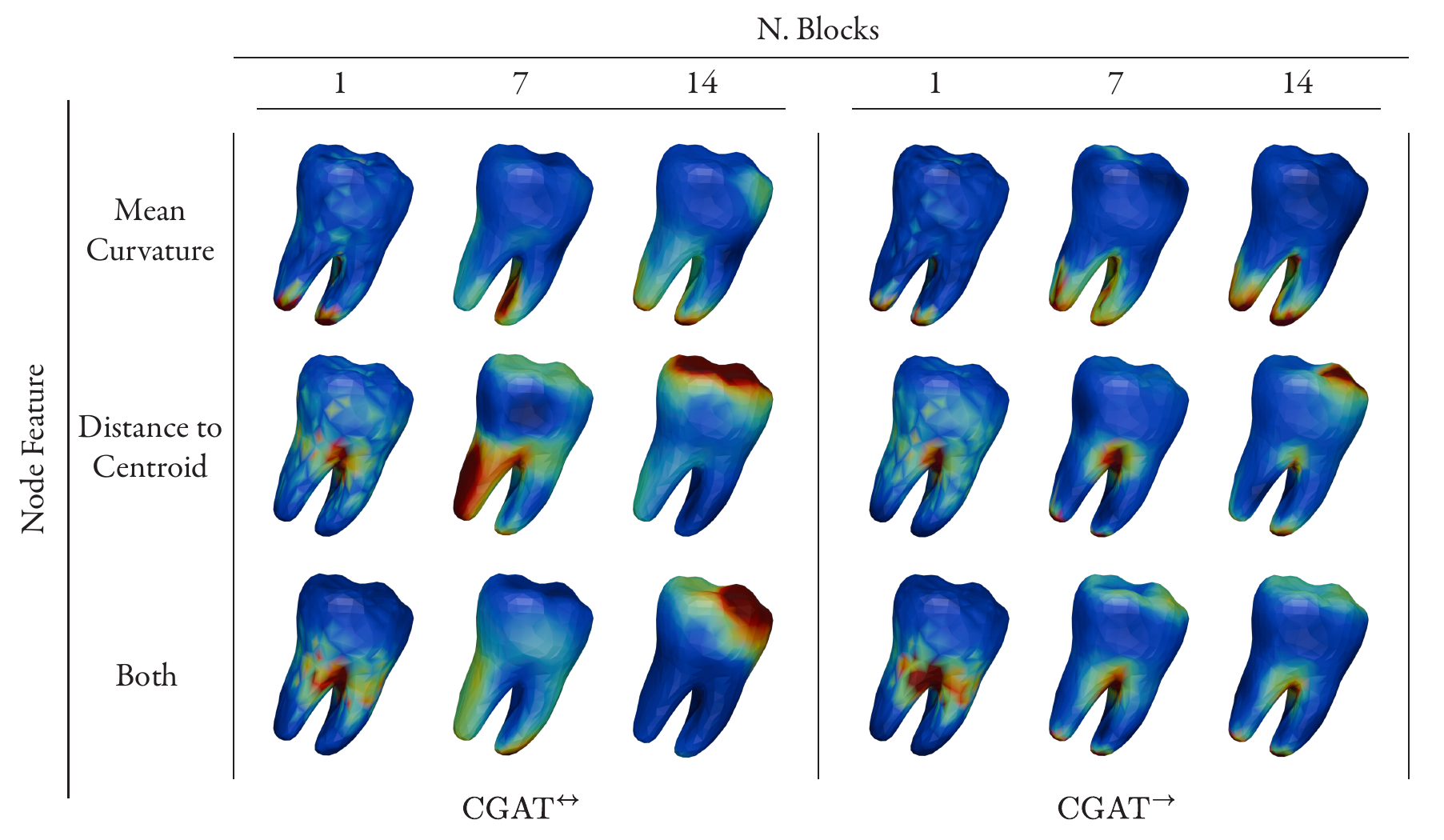}
    \caption{The attention maps from the ${}^{1,7,14}\text{CGAT}^{\{\leftrightarrow,\rightarrow\}}$ family of the tested CGAT models, visualized on a sample tooth mesh of stage H. Across both undirected $(\leftrightarrow)$ and directed $(\rightarrow)$ CLS edges, 1-block models exhibit noisy and mosaic attention, which becomes smoother and more focused in deeper models. A key difference emerges in deeper models: undirected edges tend to concentrate attention on a single anatomical region, while directed edges maintain a more distributed focus on multiple relevant areas like the root apices and furcation. This suggests that directed edges may yield more intuitive explanations, especially for deeper architectures.}
    \label{fig:attLayersUndir}
\end{figure}

Concerning the effects of number of blocks and the direction of the CLS edges on the attention maps, it is best to perform a qualitative comparison of the localization of attention for different model depths. For this comparison the models ${}^{1,7,14}\text{CGAT}_{F}^{\{\leftrightarrow, \rightarrow\}}, F\in \{curv, dist, both\}$ are selected, and the attention rollouts of these models for a sample tooth are visualized and directly compared in Fig. \ref{fig:attLayersUndir}.

The attention maps for ${}^{1,7,14}\text{CGAT}^{\{\leftrightarrow,\rightarrow\}}$ models, displayed in Fig. \ref{fig:attLayersUndir}, reveal the learned attention weights from each node to the CLS node. An immediate observation can be made on the difference of the attention maps of models with 1 block, and those with multiple blocks, with the former showing a patchy and noisy distribution, and examples of the latter being of smoother nature. This difference stems from neighborhood aggregations. With only one graph convolution operation, the attention weights of models with 1 layer are based on the initial embedding of the node features, and neighborhood aggregation does not affect the learned attention. As a result, due to the attention rollout of the first block being the attention weights of the block, the nodes show a more noisy contribution to the CLS embedding. 

Another observation that can be drawn from Fig. \ref{fig:attLayersUndir} is that when the models get deeper, regardless of the feature used, the attention maps focus on the relevant anatomical regions such as the roots and root apices, the crown of the tooth, and the furcation point of the roots. This attention pattern generally aligns with the anatomical criteria shown in Fig. \ref{fig:stageCrit}. One interesting aspect is that the attention seems to focus more on a single area in the deeper model with 14 blocks when undirected edges $(\leftrightarrow)$ are used. This effect can be explained by the receptive field of each node. More specifically, with the deeper model, all nodes are informed by their 14-hop neighbors, leading to the gathering of the information in one area through weighted aggregation. This effect is much more pronounced for the $\text{CGAT}^{\leftrightarrow}$ than for $\text{CGAT}^{\rightarrow}$, where the attentions are spread out over several regions, due to each node being a 2-hop neighbor of every other node through the CLS node. When the CLS edges are directed, this effect is somewhat rectified, as the 14-hop neighborhood is locally limited to the mesh topology instead, and as a result the attention does not focus on one single area. It can be seen that the models using directed edges $(\rightarrow)$ focus clearly on the aforementioned anatomical regions separately, even with the deepest model depicted in Fig. \ref{fig:attLayersUndir}. This effect can reduce the transparency of the deeper $\text{CGAT}^{\leftrightarrow}$ models, as the focused attentions of the deep models seem to deviate from the human understanding faster than the attentions of their directed counterparts. 

The attention maps directly reflect the effects of the chosen node features on the CLS node embedding. Because of this, the impact of the feature selection on the attention maps is highly non-trivial. This can be observed with the different regions of the sample tooth shown in Fig. \ref{fig:attLayersUndir}. With the mean curvature, regardless of the CLS edge direction, the model places more emphasis on the root apices and the root furcation, which are areas of very high convexity and concavity. With the distance to centroid feature, these regions are still attended to, but the center region where the crown and root sections join as well as the top surface of the crown show higher attentions. As the distance to the centroid is much smaller in the middle section of the teeth with elongation, the effect of the feature in this region is greater compared to the mean curvature in the same area. When the two features are combined, it can be seen that the regions of focus of models with individual singular features are both represented in the final attention. This indicates that the CGAT indeed learned the effects of both features, and was able to leverage both features in the attention mechanisms, as opposed to relying on either one only. This is also reflected in the results depicted in Fig. \ref{fig:metrics} and Table \ref{tab:results}, as the best classification performance was shown by the models using both features. This observation also corresponds with conventional clinical assessment, where both features are taken into account simultaneously by the human observer. 
Overall, the effects of the CLS edge direction seem to be more pronounced on the attention maps, compared to the effects of model depth. With a global 2-hop neighborhood enabled by undirected edges to the CLS node, in combination with the larger receptive field provided by deeper models, the attention maps seem to rely on singular regions rather than multiple relevant ones. For this, based on our observations we can state that using directed edges is more beneficial with regard to the interpretability and understandability of the attention maps, especially when the difference in the predictive performance of the two approaches is minimal. We also argue that the use of multiple features does not obscure the effects of the individual features in the final decision, and instead highlights the regions of interest covered by each feature, albeit to varying degrees. This aspect of the CGAT can help bring the explanations of the models by the way of attention maps closer to the attention patterns of human experts. 

\begin{figure}[!h]
    \centering
    \includegraphics[width=\linewidth]{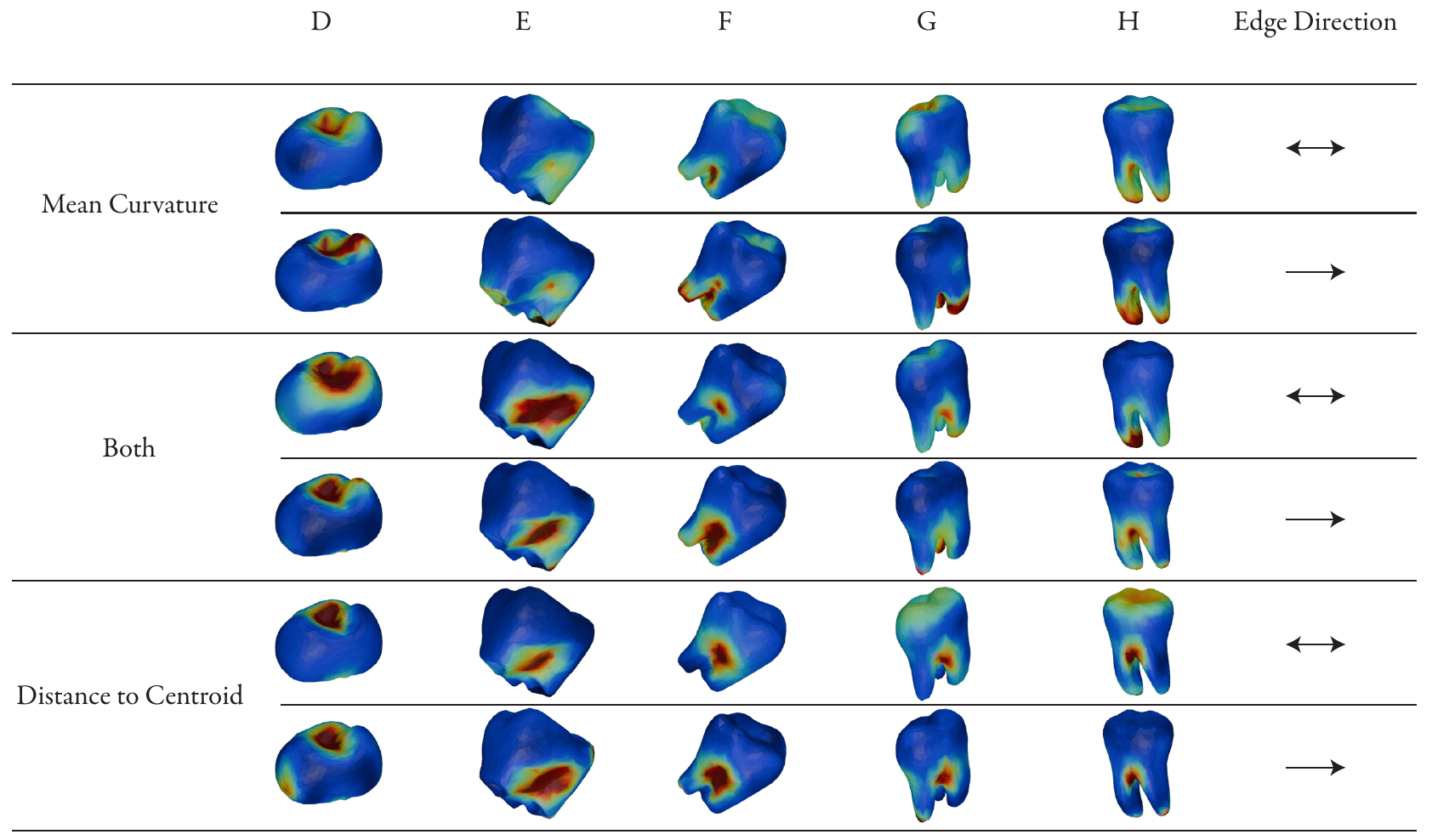}
    \caption{The attention maps of ${}^{11}\text{CGAT}^{\leftrightarrow,\rightarrow}$ for one sample tooth from all stages. At the earlier stages, the attentions are focused on the crown and the center regions of the teeth, and as the stages progress, the attention on the furcations and the roots also increase. This attention pattern is in line with the Demirjian criteria for stage assessment from Fig. \ref{fig:stageCrit}.}
    \label{fig:stageAtts}
\end{figure}   
It is also interesting to inspect how the attention evolves throughout the developmental stages. For this, we inspect the attention rollouts of ${}^{11}\text{CGAT}_{}^{\leftrightarrow}$ and ${}^{11}\text{CGAT}_{}^{\rightarrow}$ models, selected due to the metrics observed for these models in Fig. \ref{fig:metrics} indicating that this is the optimal depth for our task. An overview of the rollouts is depicted in Fig. \ref{fig:stageAtts}. In this figure, we can see an overview of the attention rollouts of the selected models. An interesting pattern of attention can be seen across all features and edge directions; across different features and edge directions, the final localizations of attentions always have significant overlap. In stage D for example, the concavity of the occlusal surface is attended to in all models. Similarly, for the later  stages, the furcation point of the roots is always considered. These results indicate that the CGAT models focus on the discriminative regions of the teeth regardless of the feature and edge direction used, since the local shape information of these regions describe the stage best. The effects of different features, then, are the amount of focus on these high-informative regions. For example, the mean curvature feature yields attentions to the crown, the furcation and the root apices in stages E and F, while the distance to centroid feature results in the attention mostly focusing to the root furcation. Similarly, for stages G and H, the mean curvature pronounces the root apices more heavily than the distance to centroid feature. We conclude that these differences stem from the distributions of the features across nodes, shown in Fig. \ref{fig:featDist}. The magnitudes of these features end up being discriminative of the developmental stage in different ranges, and as such the CGAT models learn different levels of attentions relating to them. Once again we can observe the combination of the two features in training results in attention maps that are a combination of the individually learned attentions. This effect can clearly be seen in the example for stage F. The ${}^{11}\text{CGAT}_{curv}^{\rightarrow}$ highlights the roots of the tooth, while ${}^{11}\text{CGAT}_{dist}^{\rightarrow}$ implies the beginning of the roots to be informative, and both these areas are highlighted in the attention rollout of ${}^{11}\text{CGAT}_{both}^{\rightarrow}$. Interestingly, this effect seems to be more overtly displayed for the models using directed edges to the CLS node. This once again indicates using undirected edges can result in counter-intuitive attention maps, an outcome we attribute to the global 2-hop neighboring effect. 

Overall, we conclude the CGAT architecture does indeed result in attention maps that align with the human understanding in the context of dental development, and consequently, of dental stage assessment, where the focus shifts from the crown, over the furcation, to the roots and finally the root apices.

\subsection{Performance Comparison}

To rigorously evaluate our proposed architecture, we compare its performance against several established baselines for 3D shape analysis, with the results detailed in Table \ref{tab:comparison_results}. Our proposed CGAT models demonstrate superior performance, achieving a weighted F1-Score of 0.76 and a Mean Absolute Error (MAE) of 0.25, outperforming the strongest baseline, GAT (0.67 F1-Score, 0.34 MAE). This reduction in MAE is particularly significant for an ordinal task such as dental stage assessment, as it indicates that the model's prediction errors are of a lower magnitude. Furthermore, despite a higher parameter count, CGAT remains highly efficient with inference times (7-12 ms) comparable to GAT (9 ms) and significantly faster than the computationally intensive MeshCNN (275 ms) and PointNet++ (144 ms) models.

These performance differences are rooted in the models' distinct architectural philosophies. MeshCNN operates directly on mesh topology, using task-driven edge collapse for pooling , while PointNet++ builds a hierarchy on the metric space of an unstructured point cloud via farthest point sampling. Our CGAT, and the other graph-based approaches provide a balance, leveraging topological connectivity in an abstract form. Within the graph-based methods, the performance increase over GAT can be attributed to the architectural contribution of the CLS node, which acts as a global information aggregator to create a comprehensive graph-level embedding. Ultimately, while the predictive metrics are strong, the primary advantage of CGAT remains its unique synthesis of state-of-the-art performance, computational efficiency, and the inherent ability to produce intuitive, attention-based explanations—a capability the baseline models inherently lack.

\renewcommand{\arraystretch}{1.05}
\begin{table}[h]
\centering
\caption{Performance comparison of CGAT against baseline models. We report the weighted Precision, Recall, and F1-Score, along with the Mean Absolute Error (MAE). For our proposed CGAT, we show the two best-performing variants for both directed ($\rightarrow$) and undirected ($\leftrightarrow$) CLS node edges.}
\label{tab:comparison_results}
\setlength{\tabcolsep}{4.2pt}
\begin{tabular}{@{}clcccccc@{}}
\toprule
\multicolumn{2}{c}{{Model}} & {Precision} & {Recall} & {\begin{tabular}[c]{@{}c@{}}Weighted \\ F1 Score\end{tabular}} & {MAE} & {\begin{tabular}[c]{@{}c@{}}\# of  \\ Parameters\end{tabular}} & {\begin{tabular}[c]{@{}c@{}}Inference Time \\ (ms)\end{tabular}} \\ \midrule
\multicolumn{8}{l}{\textbf{Baseline Models}} \\
\multicolumn{2}{l}{PointNet++ \cite{qi2017pointnet++}} & 0.60 & 0.63 & 0.60 & 0.48 & 1466950 & 144 \\
\multicolumn{2}{l}{MeshCNN \cite{hanocka2019meshcnn}} & 0.57 & 0.67 & 0.61 & 0.44 & 981405 & 275 \\
\multicolumn{2}{l}{GCN \cite{kipf2016semi}} & 0.66 & 0.65 & 0.64 & 0.39 & 215301 & 4 \\
\multicolumn{2}{l}{GAT \cite{velickovic2017graph}} & 0.69 & 0.65 & 0.67 & 0.34 & 1981765 & 9 \\ \midrule
\multicolumn{8}{l}{\textbf{Our Proposed Models (from the $\text{CGAT}_{both}$ family)}} \\
\multicolumn{2}{l}{${}^{6}\text{CGAT}_{both}^{\rightarrow}$} & 0.77 & 0.76 & 0.76 & 0.25 & 2499207 & 7 \\
\multicolumn{2}{l}{${}^{10}\text{CGAT}_{both}^{\rightarrow}$} & 0.72 & 0.75 & 0.73 & 0.25 & 4153479 & 9 \\
\multicolumn{2}{l}{${}^{10}\text{CGAT}_{both}^{\leftrightarrow}$} & 0.84 & 0.77 & 0.76 & 0.25 & 4153479 & 10 \\
\multicolumn{2}{l}{${}^{12}\text{CGAT}_{both}^{\leftrightarrow}$} & 0.79 & 0.78 & 0.76 & 0.27 & 4980615 & 12 \\ \bottomrule
\end{tabular}
\end{table}

\section{Conclusion}
The automation of dental stage assessment is desirable in order to eliminate the inter-observer variability, and to expedite the process. Deep learning models are the prime candidate in such applications due to their capabilities of learning complex decision patterns from the data. However, due to the high-stakes nature of dental staging in the forensic context, the opaque nature of deep learning models discourages their widespread use. In order to address this need, we introduced the CGAT architecture, which was applied to meshes of third molars derived from CBCT to determine the Demirjian stages. The CGAT model was based on the graph attention convolutions, and with its inherent attention mechanism can produce attention maps. These attention maps are produced via the attention rollout method, and are able to depict the magnitude of impact each node has on the final decision of the model, facilitated by the use of the CLS node as a global information aggregator. We evaluated the effects of different node features and the depth of the CGAT architecture on the classification performance, and the final visualization of attention maps. While all model variants were able to produce plausible attention maps, we argue that the models using directed edges to connect each node in the input meshes to the CLS node were clearer and more intuitive in their attention maps. The local mean curvature and the distance to centroid were both able to represent the local shape, from which the CGAT models were able to learn the shape structure. The combination of the two features in training not only slightly outperformed the models using individual features, but also incorporated elements from the attentions of all features into the final attention maps, which corresponds with the human expert's train of thought during conventional stage assessment. Therefore, we propose the use of $\text{CGAT}_{both}^{\rightarrow}$ as the best option to learn human-understandable attention maps.

Through the use of attention-based explanations produced by the CGAT, models can be judged with regard to their decision factors based on the attention patterns. This aspect of the CGAT architecture can increase the trust in the models and allow experts to accept or deny the model decisions. The CGAT architecture, although demonstrated on teeth meshes, is applicable to all graph data, and can perform any graph-level classification task, or regression task with small modifications to the classification section, and explain its decisions through generating attention maps. We conclude that the benefits of an architecture with increased decision transparency which can perform competitively will allow a more widespread adoption of such models with increased confidence in high-stakes settings. 

\section{Future Work}

In order to further analyze the explanation capabilities of the CGAT architecture, we argue that the merging function of the attention heads is an important avenue to explore. Many permutation-invariant merging functions exist, such as the mean and concatenation, aside from the max-pooling that we employed. Since these functions directly influence the models during training, and with that the learned attention scores, the CGAT would benefit from the exploration of this design choice. Additionally, although we only demonstrate the CGAT on dental stage classification on meshes, it is not limited to such an application only, and can be used in any graph-level task. Further applications of the CGAT models on different mesh datasets where shape information is critical, and on different graph-definable data domains such as molecular interactions or image classification, can reveal more insight into the attention-based explanations generated by CGAT.

\begin{credits}

\subsubsection{\discintname} The authors declare no conflict of interest.

\end{credits}
%
%
%
\bibliographystyle{splncs04}
\bibliography{myrefs}

\end{document}